\documentclass[]{Liebert_Author}
\usepackage[nomarkers]{endfloat}

\title{3-D Shape Control of Extensible Multi-Section Soft Continuum Robots via Visual Servoing} 

\author{Abhinav Gandhi,$^{1\ast}$ Shou-Shan Chiang,$^{1}$ Cagdas D. Onal$^{1}$, Berk Calli$^{1}$\\
{$^{1}$Robotics Engineering Department, Worcester Polytechnic Institute}\\
{100 Institute Rd, Worcester, MA 01609, USA}\\
{$^\ast$Abhinav Gandhi;}\\
{E-mail: agandhi2@wpi.edu}
}
\begin{document} 
\maketitle 
\keywords{visual servoing, shape control, soft continuum robots, whole-body control, multi-section continuum robots, soft manipulation}

\noindent \textbf{Word count:} 4,000
\begin{abstract}
In this paper, we propose a novel vision-based control algorithm for regulating the whole-body shape of extensible multi-section soft continuum manipulators. Contrary to existing vision-based control algorithms in literature that regulate the robot's end effector pose, our proposed control algorithm regulates the robot's whole-body configuration, enabling us to leverage its kinematic redundancy. Additionally, our model-based 2-1/2-D shape visual servoing provides globally stable asymptotic convergence in the robot's 3-D workspace compared to the closest works in the literature that report local-minima. Unlike existing visual servoing algorithms in literature, our approach does not require information from proprioceptive sensors, making it suitable for continuum manipulators without such capabilities. Instead, robot-state is estimated from images acquired by an external camera that observes the robot's whole-body shape and is also utilized to close the shape control loop. Traditionally, visual servoing schemes require an image of the robot at its reference pose to generate the reference features. In this work, we utilize an inverse kinematics solver to generate reference features for the desired robot configuration and do not require images of the robot at the reference. Experiments are performed on a multi-section continuum manipulator demonstrating the controller’s capability to regulate the robot’s whole-body shape while precisely positioning the robot’s end effector. Results validate our controller's ability to regulate the shape of continuum robots while demonstrating a smooth transient response and a steady state error within 1 mm. Proof of concept object manipulation experiments including stacking, pouring, and pulling tasks, are performed to demonstrate our controller’s applicability.
\end{abstract} 
\section{Introduction}
Extensible multi-section continuum manipulators possess a high number of degrees of freedom due to their modular structures. Their compliant designs allow safe interactions with delicate environments. These structural properties make them a lucrative choice for applications in highly constrained, delicate, and cluttered environments~\cite{cianchetti2018, Zhang2020a}. Specifically, they may be used in domains such as assistive and medical robotics, space robotics, and environmental robotics~\cite{liu2022review,iqbal2025continuum, kalekeyeva2025continuum}.

However, several challenges in the development and deployment of soft and Continuum Robots (CR) prevent their widespread utilization~\cite{mazzolai2022roadmap}. In this work, we develop vision-based whole-body control algorithms that enable us to leverage the workspace advantages of CRs. Since open-loop control techniques require a complete system model, that is challenging to obtain for CRs, the control accuracy is limited by the completeness of the obtained system model. Additionally, since properties of soft materials change throughout the robot's workspace and drift over time, obtaining accurate model parameters is not possible. Thus open-loop control of these robots is not feasible. In such cases, closed-loop controllers are utilized for overcoming modeling uncertainty~\cite{della_santina_model-based_2023}. However, these techniques require state information about the robot's body, that are typically obtained from proprioceptive sensors. Although various proprioceptive sensing approaches have been explored in soft robotics literature~\cite{russo2023continuum,roesthuis2016steering,monet2020high}, they remain challenging to reliably manufacture and integrate into novel CRs. Keeping in mind these challenges, we propose a novel control algorithm that utilizes purely visual cues for control loop closure without on-board proprioceptive sensing capabilities.

Vision-based control algorithms in the literature can be classified into two broad categories based on how the control error is defined: Image-Based Visual Servoing (IBVS), where the error is defined in the image-space, and Position-Based Visual Servoing (PBVS), where the error is defined in cartesian-space~\cite{hutchinson1996tutorial}. Although these approaches have been widely utilized in robot control literature~\cite{nazari2022visual,yip2014model}, they have their drawbacks. Position-based approaches like~\cite{dong2015position, wilson1996relative} require 3-D object models with reference to which the positioning task is carried out. Although image-based approaches do not require object models, they do not provide global convergence guarantees~\cite{kragic2002survey}. Existing works in IBVS have reported local minima in the robot's workspace~\cite{gandhi2022SkeletonbasedAdaptiveVisual, Xu2022}. To overcome these drawbacks, a 2-1/2-D visual servoing approach is proposed in~\cite{malis19992} to control the robot's end effector pose. While this method combines the advantage of IBVS and PBVS, it only controls the end effector pose and not the robot's whole body-shape which can lead to configuration uncertainty for CRs. In contrast, we utilize the 2-1/2-D visual servoing framework to control the whole-body configuration of CRs thus enabling us to leverage their kinematic redundancy. To obtain visual feedback, existing visual servoing formulations in robotics literature utilize two types of camera setups: the eye-in-hand setup, where the camera is mounted to the robot's end effector, and the eye-to-hand setup, where an external camera observes the robot's end effector. However, these formulations only control the robot's end effector and fail to leverage their kinematic redundancy. To control the whole-body pose of the robot, we introduce an eye-to-body camera setup where an external camera observes features along the robot's body. Compared to the existing eye-to-hand approaches, that also utilize an external camera, our approach enables configuration control of CRs enabling us to leverage their kinematic redundancy.

Prior work in vision-based control of robots, including image-based, position-based, and 2-1/2-D visual servoing, focuses on controlling the robot's end effector pose~\cite{hutchinson1996tutorial,nazari2022visual, yip2014model,kamtikar2022visual,dong2015position, wilson1996relative, malis19992,fang2002adaptive} thus failing to leverage any configuration redundancies. Inspired by shape control approaches in deformable object manipulation literature~\cite{Navarro-Alarcon2018}, our prior work in~\cite{gandhi-clothoids, gandhi-growtoshape} and other work in~\cite{Xu2022} explores the idea of controlling the configuration of the robot using IBVS. In~\cite{Xu2022} the authors utilize Bezier curves to control the shape of an 8 DoF tendon-driven soft robotic manipulator with fixed length. However, they report existence of local minima in the workspace when driving the robots to S-shape configurations. Local minima are also reported in our prior work~\cite{gandhi2022SkeletonbasedAdaptiveVisual}, where we utilize B-splines to model and control the shape of a rigid manipulator. To overcome the local minima, authors in~\cite{Xu2022} propose utilizing additional image features to drive the robot to S-shape configurations. Since this approach requires specially designed image feature sets for driving the robot to different types of shape references in the image, it is not quite suitable for practical applications. Since IBVS approaches do not provide global convergence guarantees, local minima are not uncommon in such systems. To enhance the utility of these image-based shape control techniques, we investigated clothoids as shape representations to control the configuration of a two-section variable length continuum manipulator in~\cite{gandhi-clothoids}. In this approach, points along the clothoid and their curvatures are utilized as image features for control. We demonstrated that utilizing curvature information from the clothoid along with point-features successfully overcomes local minima. However, clothoids are 2-D curves which limits control to a plane. Furthermore, the parametric representations chosen in both~\cite{gandhi-clothoids, Xu2022} require additional consideration to scale to multiple sections. In our latest work~\cite{gandhi-growtoshape}, we utilized the common piecewise constant curvature arc geometry to estimate and control the robot's shape. The proposed approach allows us to scale control to multi-section CRs but is limited to 2-D and only controls the sections sequentially to grow the robot. All these existing vision-based shape control algorithms in soft robotics literature~\cite{gandhi-clothoids, Xu2022,gandhi-growtoshape} utilize model-free adaptive visual servoing frameworks. While these methods are advantageous when the system model is entirely unavailable, their transient response is generally unpredictable and may be unsuitable for performing manipulation tasks. Overall, the literature exploring vision-based shape control of multi-section CRs is limited.

In this work, we propose a novel model-based 2-1/2-D visual servoing algorithm to control the whole-body shape of multi-section extensible continuum manipulators in 3-D. To achieve this, we design a novel robot-shape Jacobian which enables control of the robot's whole-body shape. The proposed shape Jacobian is modular, and its general form can be scaled for multi-section CRs as needed. The robot's state is estimated by utilizing purely visual information thus eliminating the need for proprioceptive sensors. This simplifies the on-board hardware for CRs. With our eye-to-body setup, we achieve whole-body control of multi-section continuum manipulators enabling us to leverage their workspace advantages in cluttered environments. We also utilize our research team's prior work~\cite{chiang-amorph}, which develops an optimization-based inverse kinematics solver, to intuitively select reference shapes for the robot in the acquired camera image.
\section{Materials and Methods}\label{sec:methods}
\begin{figure}[ht]
    \centering
    \includegraphics[width=\textwidth]{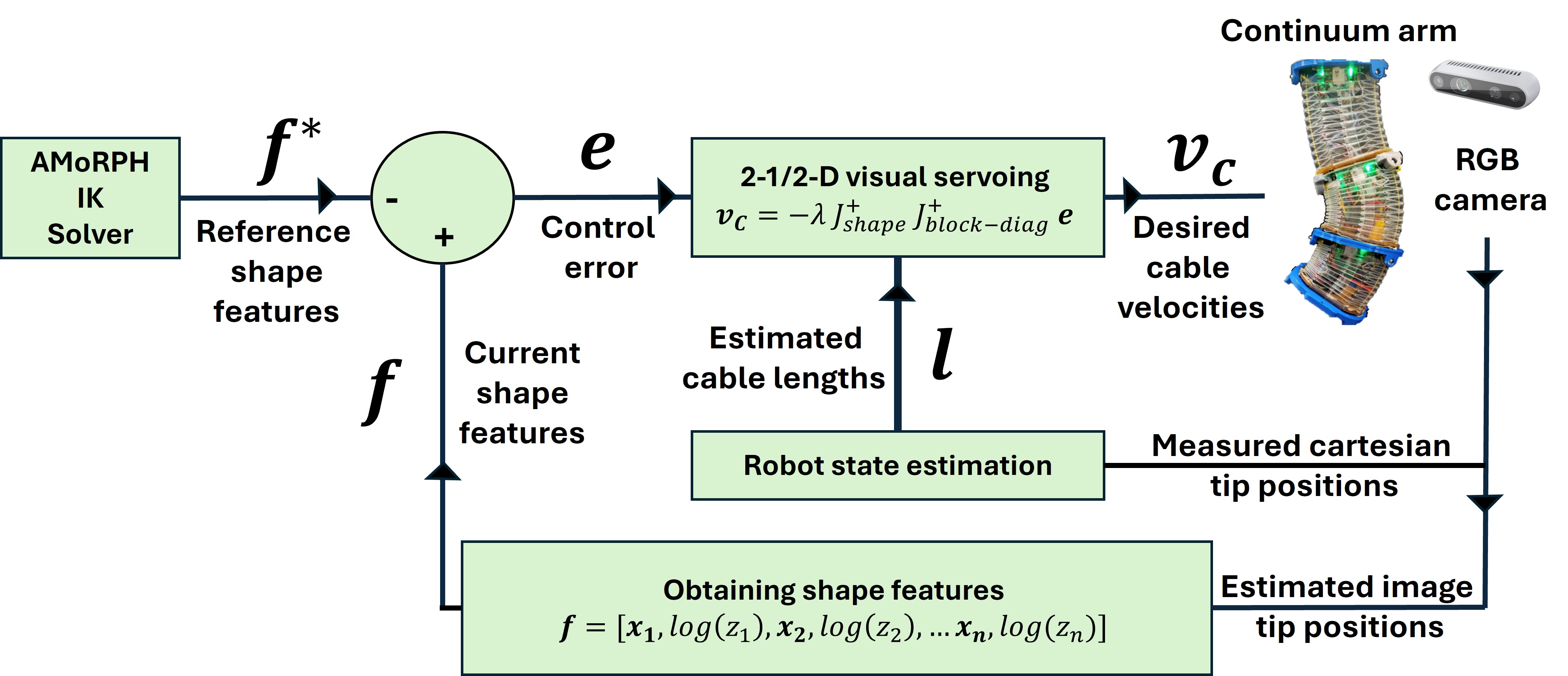}
    \caption{Controller block-diagram.}
    \label{fig:block-diag}
\end{figure}

The schematic (Fig.~\ref{fig:block-diag}) provides an outline of our algorithm and we discuss the details in the remainder of this section. We first introduce the CR and its kinematic model utilized in our study. Afterwards, we present our novel shape visual servoing algorithm.
\subsection{Extensible Continuum Arm}\label{sec:orca}
The proposed vision-based whole-body shape control algorithm, is applicable to a broad range of fixed length and extensible CRs. In this study, we utilize an extensible CR manipulator, shown in (Fig.~\ref{fig:exp-setup}), that consists of origami inspired modular continuum sections~\cite{santoso2017DesignAnalysisOrigami}. Each section of the CR is driven by $3$ individually actuated tendons. This tendon-driven mechanism provides each section with $3$ controllable degrees of freedom: including bending in $2$ axis and length modulation. The sections can contract to a minimal length of $80 mm$ and expand to a maximal length of $200 mm$. They are constructed from PET sheets which are folded using the well-known Yoshimura crease pattern. Such tendon actuated designs along with a lack of proprioceptive sensing are a common feature of CRs in soft robotics literature. Our proposed vision-based shape control scheme removes the reliance on proprioceptive sensing for control by utilizing purely visual cues to estimate and control the CR's whole-body shape. 
\subsection{Multi-Section Continuum Manipulator Kinematics: PUP Model}\label{sec:continuum-kinematics}
\begin{figure}[htbp]
    \centering
    \includegraphics[width=0.5\textwidth]{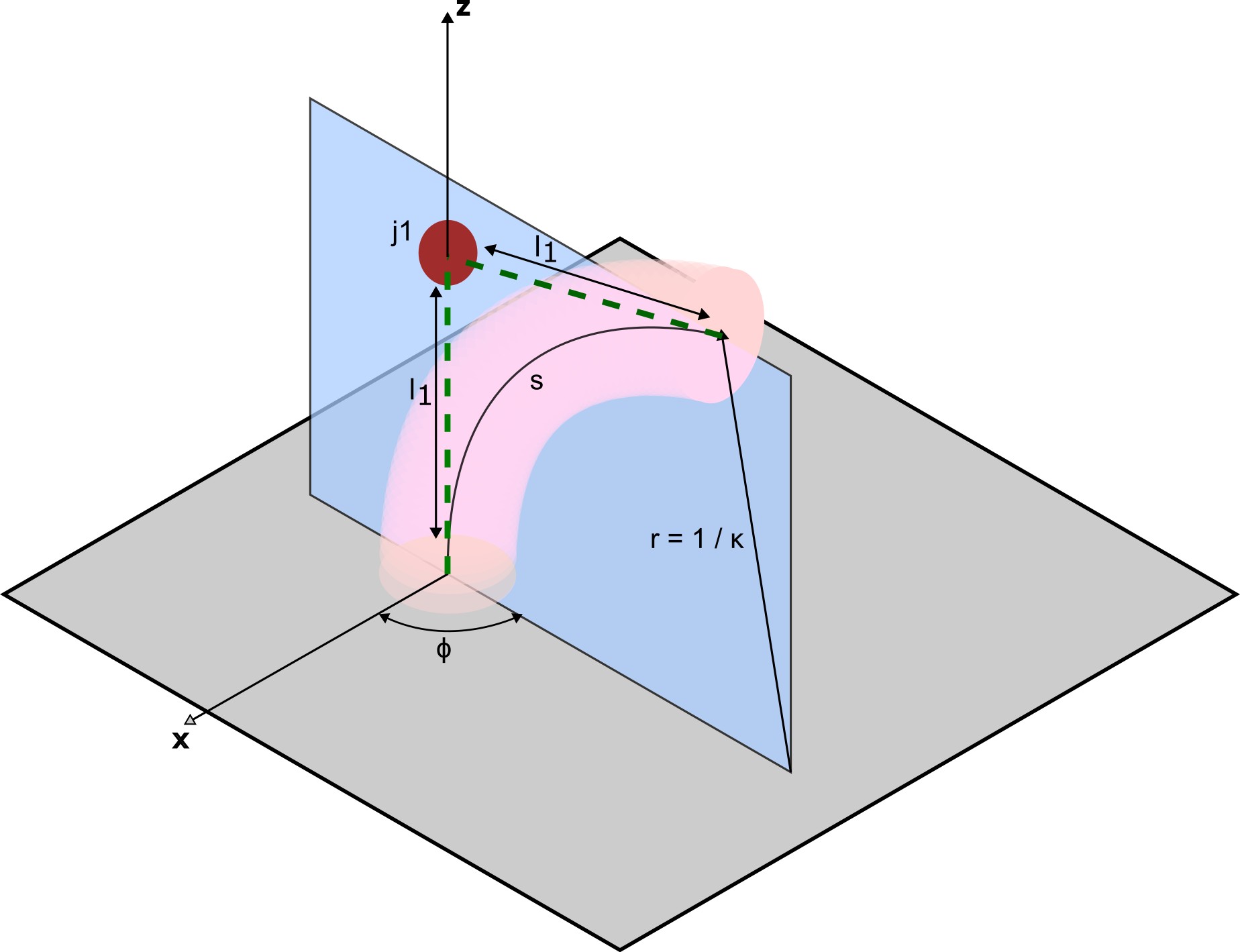}
    \caption{PUP model}
    \label{fig:pup-model}
\end{figure}
To control the robot's whole-body configuration, we parametrize its shape with Piecewise Constant Curvature (PCC) arcs, such that each continuum section of the robot is represented by an arc segment~\cite{webster-jones-review}. The PCC model serves as a robot independent mapping between the arc parameters and the robot's tip. Furthermore, the arc parameters can be mapped to the robot's cable lengths by a robot dependent mapping. We utilize a rigid body approximate model (Fig.~\ref{fig:pup-model}) to obtain the robot dependent mapping. A similar approach utilizing a virtual joint model has been adopted in~\cite{zhang2018fabrikc}. However, the virtual joint model, utilized in~\cite{zhang2018fabrikc}, models each continuum section as a pair of virtual links of fixed length connected to a virtual joint. Since we develop this model for the more general case of variable arc-length, as seen in extensible continuum manipulators, we utilize a pair of virtual prismatic joints, with variable length $l_1$, instead of the fixed length virtual links. To ensure model compatibility with the constant curvature constraint, the prismatic joint pairs always maintain the same length. The universal joint $j_1$ is located at the intersection of the prismatic joints. We refer to this approximation as the Prismatic-Universal-Prismatic (PUP) model. Using the 3 DoF PUP model, we now have a one-on-one mapping between the kinematics of the CR and its parametric shape space represented by: arc-length ($s$), curvature ($\kappa$), and bending-direction ($\phi$). The Denavit-Hartenberg parameters for our virtual links representation of the PUP model, for a single continuum section, are shown in Table~\ref{tab:D-H-params}. A homogeneous transformation $T_{i}$ is derived between the base and tip of each continuum section by following the D-H method. The transformations $T_i$ are multiplied in succession to obtain the homogeneous transformation between the robot's base and end effector, resulting in the robot's forward kinematics. The robot Jacobian $J_{robot}$ is derived by application of chain rule. In Section~\ref{ssec:shape-Jacobian}, we utilize this robot Jacobian as a building block for our novel robot-shape Jacobian.
\begin{table}[ht]
\centering
\begin{threeparttable}
    \caption{D-H Parameters for PUP linkage}
    \label{tab:D-H-params}
    \begin{tabular}{c||c c c c}
    \toprule
    Link & a & $\alpha$ & d & $\theta$  \\
    \midrule
     1 & $0$ & $0$ & $0$ & $q_1$ \\
     2 & $0$ & $0$ & $q_2$ & $0$ \\
     3 & $0$ & $-\pi/2$ & $0$ & $q_3$\\
     4 & $0$ & $\pi/2$ & $q_4$ & 0\\
     5 & $0$ & $0$ & $0$ & $q_5$\\
    \bottomrule
    \end{tabular}
\end{threeparttable}
\end{table}
\subsection{2-1/2-D Shape Visual Servoing}\label{ssec:visual-servoing}
In robot control literature, 2-1/2-D visual servoing formulations have been successfully applied to control the robot's end effector pose. These existing approaches track features, $\mathbf{f}$, at the robot's end effector. The tracked features consist of a combination of feature locations in the image and their associated depths. To drive the robot to a desired position in its workspace, a control error signal, $\mathbf{e}$, is defined between the current features, $\mathbf{f}$, and the desired features, $\mathbf{f}^*$, as shown in Equation~\eqref{eq:error}.
\begin{equation}
    \label{eq:error}
    \mathbf{e} = \mathbf{f} - \mathbf{f}^*
\end{equation}
The control error, $\mathbf{e}$, is utilized in a visual servo law~\cite{hutchinson1996tutorial} that is designed to exponentially minimize the error between the current and desired features, and as a result drives the robot's end effector from its current pose to the desired pose. The controller utilizes an image Jacobian and a robot Jacobian to generate desired control inputs for the robot's actuators. The image Jacobian, $J_{img}$, projects the image feature error as desired cartesian feature velocities as seen in Equation~\eqref{eq:image-mapping}.
\begin{equation}\label{eq:image-mapping}
    \dot{\mathbf{f}}_{cart} = J_{img}^{+} \mathbf{e}
\end{equation}
The image Jacobian, $J_1$ for a point feature $\mathbf{P}_1$ is shown in Equation~\eqref{eq:image-jacobian}. Here, $x_1$ and $y_1$ are the pixel co-ordinates of $\mathbf{P}_1$ in the image and $z_1$ is its estimated depth. For multiple features $\mathbf{P}_1, \mathbf{P}_2, \dots, \mathbf{P}_n$, additional rows are appended to the image Jacobian as shown in Equation~\eqref{eq:image-jacobian-concat}.
\begin{equation} \label{eq:image-jacobian}
    J_{img} = J_{1} = \begin{bmatrix}
        \lambda & 0 & -x_1/z_1\\
        0 & 1/\lambda & -y_1/z_1\\
        0 & 0 & -1\\
    \end{bmatrix}
\end{equation}
\begin{equation}\label{eq:image-jacobian-concat}
    J_{img} = \begin{bmatrix}
        J_{1} & J_{2} & \dots & J_{n}
    \end{bmatrix}^T
\end{equation}
The cartesian feature velocities, $\dot{\mathbf{f}}_{cart}$, are mapped to a set of desired cable velocities for the robot using the inverse robot Jacobian, $J_{robot}$, as seen in Equation~\eqref{eq:cable-vels}. 
\begin{equation}\label{eq:cable-vels}
    \mathbf{v}_{cable} = J_{robot}^{+} \dot{\mathbf{f}}_{cart}
\end{equation}
Detailed derivations of the control law and 2-1/2-D image Jacobian can be found in~\cite{hutchinson1996tutorial, malis19992} respectively. Since redundant robots have multiple valid robot configurations for an end effector pose, the controller in Equation~\eqref{eq:cable-vels} is only sufficient for regulating the robot's end effector and does not guarantee convergence at the desired configuration.

To control the whole-body configuration of the robot we observe shape features, as discussed in Section~\ref{ssec:shape-features}, along the robot's entire body instead of tracking features purely at the robot's end effector. Additionally, we introduce a novel block-diagonal image Jacobian, $J_{img'}$, that is discussed in Section~\ref{ssec:diag-img-jacobian} and a robot-shape Jacobian, $J_{shape}$, that is discussed in Section~\ref{ssec:shape-Jacobian}. These replace the conventional image and robot Jacobian discussed before and we arrive at our modified visual servoing law that is shown in Equation~\eqref{eq:control-law}, where $\lambda$ is the visual servo gain.
\begin{equation}\label{eq:control-law}
    \mathbf{v}_{cable} = - \lambda J_{shape}^+ J_{img'}^{+} \mathbf{e}
\end{equation}
\subsection{Shape Features}\label{ssec:shape-features}
As mentioned before, we observe features along the robot's body and utilize them as our image features for controlling the robot's whole-body pose. Our image feature vector, $\mathbf{f}$, is defined in a hybrid format and comprises of points along the robot's body, $\mathbf{x}_i$, and a normalized depth, $log(Z_i)$, at these points. The image feature vector, for a robot with $N$-sections, is shown in Equation~\eqref{eq:image-features}. Our choice of features for visual servoing encapsulates information about the robot's entire shape thus enabling us to control its whole-body configuration and leverage its kinematic redundancy. 
\begin{equation}
    \label{eq:image-features}
    \mathbf{f} = [\mathbf{x}_1, log(z_1), \mathbf{x}_2, log(z_2),\dots, \mathbf{x_N}, log(z_N)]
\end{equation}
\subsection{Block-diagonal Image Jacobian}\label{ssec:diag-img-jacobian}
As discussed in Section~\ref{ssec:visual-servoing}, the image Jacobian projects the image feature error to cartesian feature velocities. Traditionally, for 2-1/2-D visual servoing, the image Jacobian for a single point feature is a matrix of dimension $3 X 3$ as shown in Equation~\eqref{eq:image-jacobian}. For additional image features, the image Jacobian contains additional rows concatenated vertically, as shown in Equation~\eqref{eq:image-jacobian-concat}. This mapping asserts a velocity equal to the camera velocity to all the image features and ensures end effector pose convergence. However, since our formulation observes features along the entire robot body, we require a mapping that decouples the feature velocities from the robot's end effector. Our proposed block-diagonal formulation for the image Jacobian as seen in Equation~\eqref{eq:block-diag-jacobian} decouples the cartesian velocities for all the point features enabling configuration level control of the robot.
\begin{equation}\label{eq:block-diag-jacobian}
    J_{img'} = \begin{bmatrix}
J_1 & 0 & \cdots & 0 \\
0 & J_2 & \cdots & 0 \\
\vdots & \vdots  & \ddots & \vdots \\
0 & 0 & \cdots & J_N
\end{bmatrix}
\end{equation}
\subsection{Robot-shape Jacobian}\label{ssec:shape-Jacobian}
Since we control the whole-body configuration of the robot, we design a robot-shape Jacobian denoted by $J_{shape}$. The robot-shape Jacobian maps robot's cable velocities to tip velocities of each module of the robot and enables configuration control of the CR. To clarify further, the robot Jacobian $J_{robot}$ shown in Section~\ref{sec:continuum-kinematics} maps cable velocities of the robot to its end effector velocities. Consequentially, its inverse mapping would lead to configuration uncertainty in case of redundant robots. Our proposed robot-shape Jacobian overcomes this configuration uncertainty by utilizing the tip velocities of each section of the robot as shown in Equation~\eqref{eq:shape-Jacobian}, for a three-section robot, where $J_{ri}$ is the robot Jacobian for $i$ continuum sections and has a dimension $3X3i$. Notably, our proposed robot-shape Jacobian formulation is modular and can be applied to CRs with any number of sections. This shape Jacobian is applied in the visual servo law, to enable whole-body control of CRs, as discussed in the following Section~\ref{ssec:visual-servoing}.
\begin{equation} \label{eq:shape-Jacobian}
    \begin{bmatrix}
    \mathbf{v}_{tip1}\\ \mathbf{v}_{tip2}\\ \mathbf{v}_{tip3}    
    \end{bmatrix}
     = 
    \begin{bmatrix}
        J_{r1 [3X3]} & [0]_{3X6}\\
        J_{r2 [3X6]} & [0]_{3X3}\\
                J_{r3 [3X9]} 
    \end{bmatrix}
    \begin{bmatrix}
    \dot{\mathbf{l}}_1\\
    \dot{\mathbf{l}}_2\\
    \dot{\mathbf{l}}_3        
    \end{bmatrix}
\end{equation}
\subsection{Robot State Estimation}
Utilizing the PUP model discussed in Section~\ref{sec:continuum-kinematics} we can represent the robot's state with robot dependent actuator space parameters, $\{l_{i1}, l_{i2}, l_{i3}\}$, or robot independent PCC arc parameters, $\{s_i, \kappa_i, \phi_i\}$, for each continuum section. In our method, we estimate the robot's state directly from the image by fitting PCC arcs to the robot's observed shape. The robot independent state is converted into the robot dependent mapping as discussed in~\ref{sec:continuum-kinematics} utilizing the closed form inverse mapping~\cite{jones2006kinematics} shown in~\ref{eq:cable-len-estimation}. The obtained cable lengths representing the robot's state are substituted in the robot-shape Jacobian at each cycle of the control loop.
\begin{equation}\label{eq:cable-len-estimation}
\begin{bmatrix}
    l_1 \\ l_2 \\ l_3 
\end{bmatrix} = 
\begin{bmatrix}
    s(1-\kappa d sin\phi)\\ s(1+\kappa d sin(\pi/3 + \phi))\\s(1-\kappa d cos(\pi/6 + \phi)
\end{bmatrix}
\end{equation}
\subsection{Reference Shape Generation}
Traditionally, visual servoing algorithms require target feature locations in the image to servo the robot. Existing implementations in the literature obtain target features by taking a snapshot of the robot at the reference. Such an approach limits practical application. In this paper, we utilize our research team’s prior work (AMoRPH)~\cite{chiang-amorph} to intuitively generate reference shapes for multi-section extensible CRs. To provide a reference PCC shape, AMoRPH requires the desired end effector pose in the image which can be intuitively selected using a mouse pointer and the robot’s base pose. Using these, a geometric optimization provides a solution for a two-segment PCC arc. Additional arc segments are inserted using an iterative approach to match the number of continuum sections. The generated PCC references can be optimized by balancing curvature or arc lengths of the continuum sections. Additionally, references can be generated to avoid known obstacles in the environment. Reference shape features for visual servoing are obtained by selecting the points at the intersection of each arc segment. Note that our implementation of AMoRPH is limited to the image plane. Integrating 3-D AMoRPH is out of scope for this work since the focus of this work is the novel shape control algorithm.
\section{Experimental Results}\label{sec:results}
Experiments are performed to evaluate the control performance and test the reliability of the shape control. Additionally, manipulation demonstrations are performed to showcase the system’s strengths. The experiments are performed on a two-section and three-section setup of the CR in Section~\ref{sec:orca}. Details about the experiment setup and results obtained are discussed below.
\begin{figure}[htbp]
    \centering
    \includegraphics[width=0.6\textwidth]{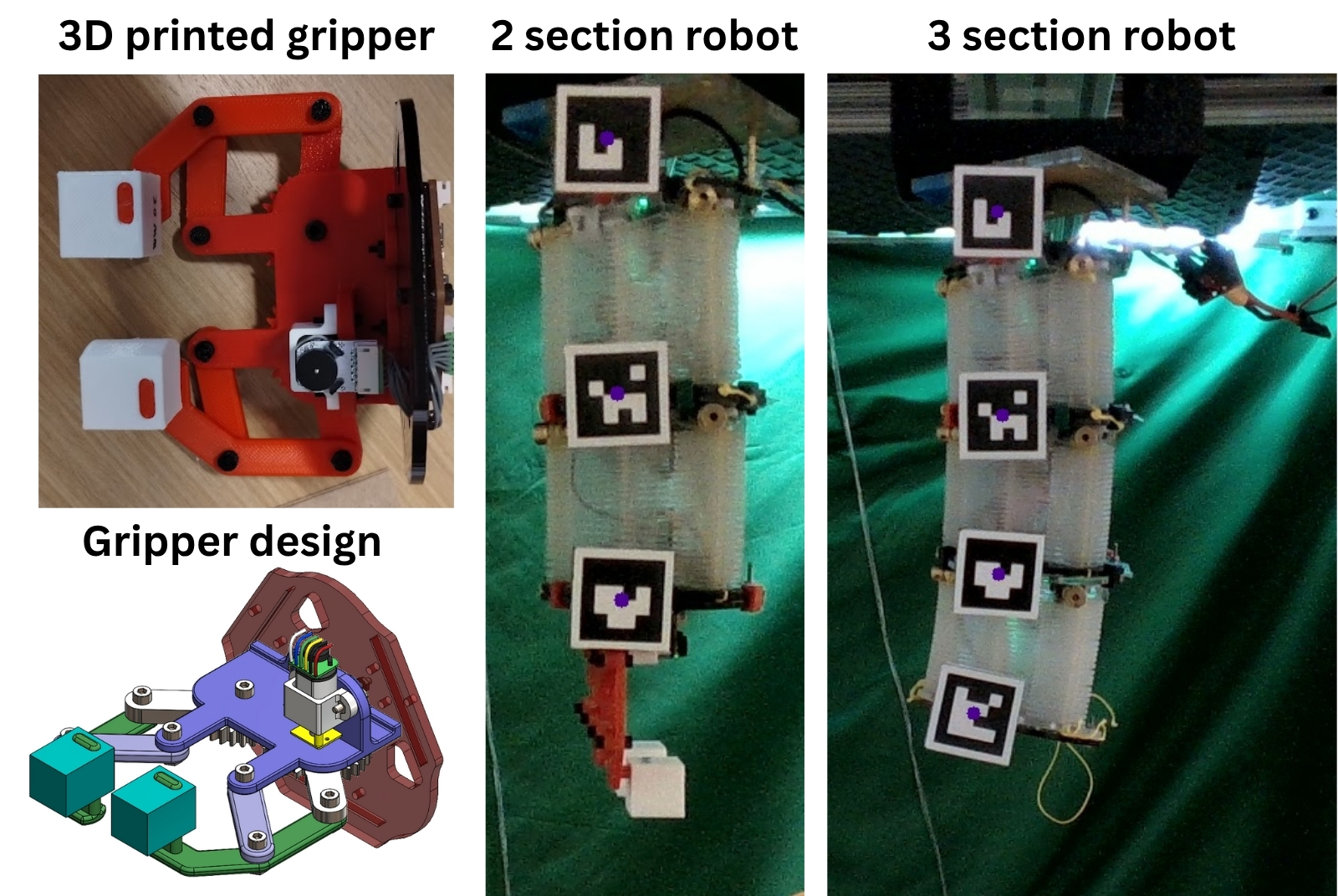}
    \caption[Experimental setup]{Figure demonstrating the experimental setup including the 3-D printed gripper and a rendering of its CAD model, a two-section origami CR with the gripper attached to its end effector, and a three-section origami CR.}
    \label{fig:exp-setup}
\end{figure}
\subsection{Experimental Setup}\label{sec:experimental-setup}
We utilize up to three sections of the origami continuum manipulator~\cite{santoso2017DesignAnalysisOrigami} to perform our experiments. The manipulator is mounted to a horizontal bar. Its mounting position is chosen to ensure an unobstructed camera view and access to a horizontal surface for performing manipulation demonstrations. An externally mounted camera (Intel RealSense d435i) captures RGB images ($1280 X 720$ pixel resolution) of the robot's whole-body configuration. Shape feature locations are estimated by tracking ArUco markers placed on the tips of the continuum sections. Note that we minimize the use of markers by only utilizing minimal information such as the feature locations and depth. We \textbf{do not} track the 6 DoF pose of the markers. Alternatively, shape features can be tracked with a higher resolution RGB-D camera without specialized markers but that is outside the scope of this work. To test our controller’s ability to manipulate objects, we utilize a lightweight parallel jaw gripper mechanism.

At the beginning of each experiment, all the sections of the robot are retracted to their minimal length to ensure uniformity across experiments. The control frequency for our experiments is set to $10 Hz$. An image of the experimental setup, including the continuum manipulator and the 3-D printed gripper is shown in Fig.~\ref{fig:exp-setup}. The remainder of this section describes the experiments conducted and discusses their outcomes.
\subsection{Evaluation of 3-D Shape Control}
We conduct a total of $14$ experiments on a two-section and three-section continuum manipulator to evaluate the \textit{transient response and steady-state convergence} of our proposed whole-body shape control algorithm. The controller is tuned on a sample shape reference, to ensure no overshoot and a steady-state error of less than $5 \%$. Our results demonstrate that the controller successfully achieves desired 3-D shapes and end effector pose while maintaining smooth trajectories in the image along with minimal overshoot and low steady-state error. Experiments are presented in supplementary videos S1 and S2.
\begin{figure}[ht]
    \centering
    \includegraphics[width=\textwidth]{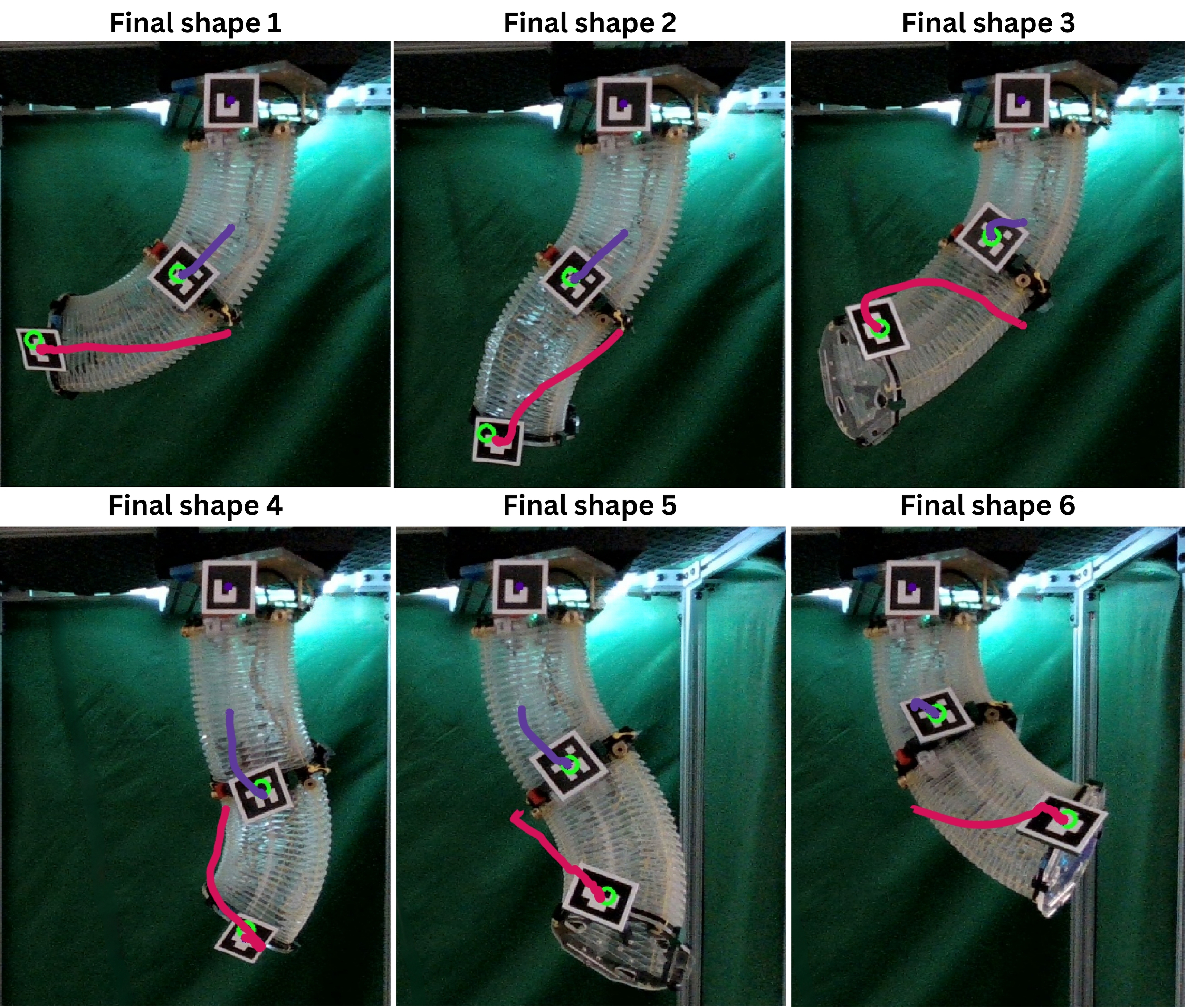}
    \caption{The final achieved shape for 6 shape control experiments performed with two sections of the continuum manipulator are shown. Feature trajectories are depicted by the purple and red curves shown in the images. The reference feature locations in the image are shown by the green circles.}
    \label{fig:two-module-results}
\end{figure}
\begin{figure}[ht]
    \centering
    \includegraphics[width=\textwidth]{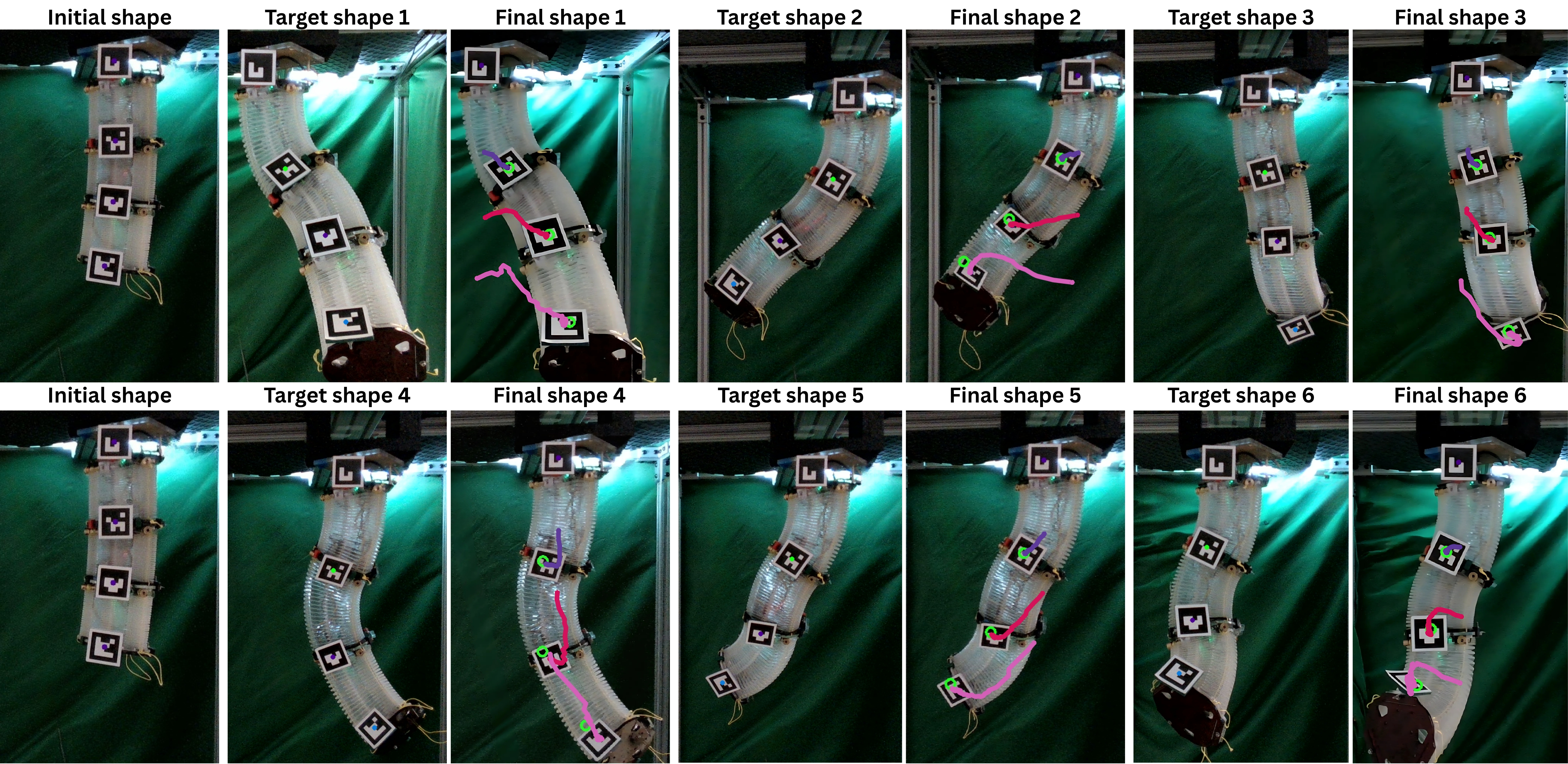}
    \caption{The final achieved shape for 6 shape control experiments performed with three sections of the continuum manipulator are shown. Feature trajectories are depicted by the purple, red, and pink curves shown in the images. The reference feature locations in the image are shown by the green circles.}
    \label{fig:three-module-results}
\end{figure}
\subsubsection{Controller convergence}
To evaluate the convergence of our vision-based shape control, we choose $6$ randomly generated shape references each for the two-section and three-section manipulator. Steady-state errors for task-space convergence and configuration-space convergence are reported in Table~\ref{tab:steady-state}, averaged over the $6$ experiments. To measure the steady-state errors an $L2$ norm metric is used. In the configuration-space, the norm is reported for the cumulative error between the observed and desired shape feature vector discussed in Section~\ref{ssec:shape-features}. Whereas, to evaluate the task-space convergence we report the $L2$ norm of the observed error for only the shape feature associated with the robot's end effector. Our results show depth positioning error of less than $5 mm$ and image plane positioning errors of less than $15 px$ in the task-space for the two-section robot indicating a high accuracy of control. Noticeably, the steady-state error for the three-section robot is larger. This is due to the compounding error over each section of the robot and the non-linear effects of the system's dynamics which become more pronounced when utilizing three sections.
\begin{table}[ht]
  \centering
  \caption{Steady-state errors}
  \label{tab:steady-state}
  \begin{tabular}{lcc}
    \hline
    \textbf{\# sections} & \multicolumn{2}{c}{\textbf{Steady-state error (pixels / mm)}} \\
    \cline{2-3}
     & \begin{tabular}[c]{@{}c@{}}Task-Space\end{tabular} & \begin{tabular}[c]{@{}c@{}}Configuration-Space\end{tabular} \\
    \hline
    & Image-Space / Depth & Image-Space / Depth \\
    \hline
    $2$ & $9.5\pm4.7$ / $3.6\pm1.2$ & $10.2\pm4.9$ / $3.7\pm1.2$ \\
    3 & $14.3\pm6.9$ / $6.7\pm4.5$ & $16.8\pm8.6$ / $7.5\pm5.5$  \\
    \hline
  \end{tabular}
\end{table}
\subsubsection{Transient response}
In addition to analyzing the steady-state performance of our proposed controller, we also report the performance of our system during the transient stage. The transient stage metrics include: \textit{system rise-time and system settling-time}. 

We define two criteria to compute system rise-time and settling-time. A stringent criterion is applied to the two-section system while a relaxed one is applied to the three-section system. The relaxed criteria are utilized for the three-section robot since it has a larger steady-state error. The rise-time is computed as the time taken for the error norm to decay from $90\%$ to $10\%$ of the total error at the beginning of the experiment and from $80\%$ to $20\%$ of the total error for the relaxed criterion. We consider the system to be settled once the error norm falls below $5\%$ and $10\%$ of the total error recorded at the experiment start time for both setups respectively. Results are averaged over $6$ experiments each for the two setups and reported in Table~\ref{tab:transient-response}. For the two-section case, we report metrics for both the stringent and relaxed criterion. We can confirm that the smaller continuum arm has a shorter rise-time and settling-time compared to the three-section arm for the same criteria. Anecdotally, the feature trajectories shown in Fig.~\ref{fig:two-module-results} and Fig.~\ref{fig:three-module-results} indicate a smooth and near optimal transient performance in the image. Such predictable transient response is beneficial for applications that require the robot to operate in cluttered environments.
\begin{table}[ht]
\centering
\begin{threeparttable}
    \caption{Transient response}
    \label{tab:transient-response}
    \begin{tabular}{c c c}
    \toprule
    $\mathbf{\#}$ \textbf{sections}&\textbf{System}&\textbf{System}\\
     &\textbf{rise-time} (s)&\textbf{settling-time} (s)\\
    \midrule
     2 (stringent) & $4.8\pm1.9$ & $12.5\pm1.1$\\
     2 (relaxed) & $1.9\pm0.9$ & $5.1\pm2.1$\\
     3 & $2.75\pm1.6$ & $8.2\pm3.1$\\
    \bottomrule
    \end{tabular}
\end{threeparttable}
\end{table}
\subsection{Object manipulation}
To demonstrate the applicability and reliability of our controller, three object manipulation tasks - cup stacking, rice pouring, and drawer pulling - are performed. Additionally, a reliability test is performed by running 5 consecutive trials of the cup stacking task. For each of these object manipulation tasks, we utilize the parallel jaw gripper (Fig.~\ref{fig:exp-setup}). The cup stacking and rice pouring tasks are performed with the two-section robot setup. Since the drawer pulling task requires a larger workspace, we use the three-section robot.
\subsubsection{Cup stacking}
\begin{figure}[htbp]
    \centering
    \includegraphics[width=\textwidth]{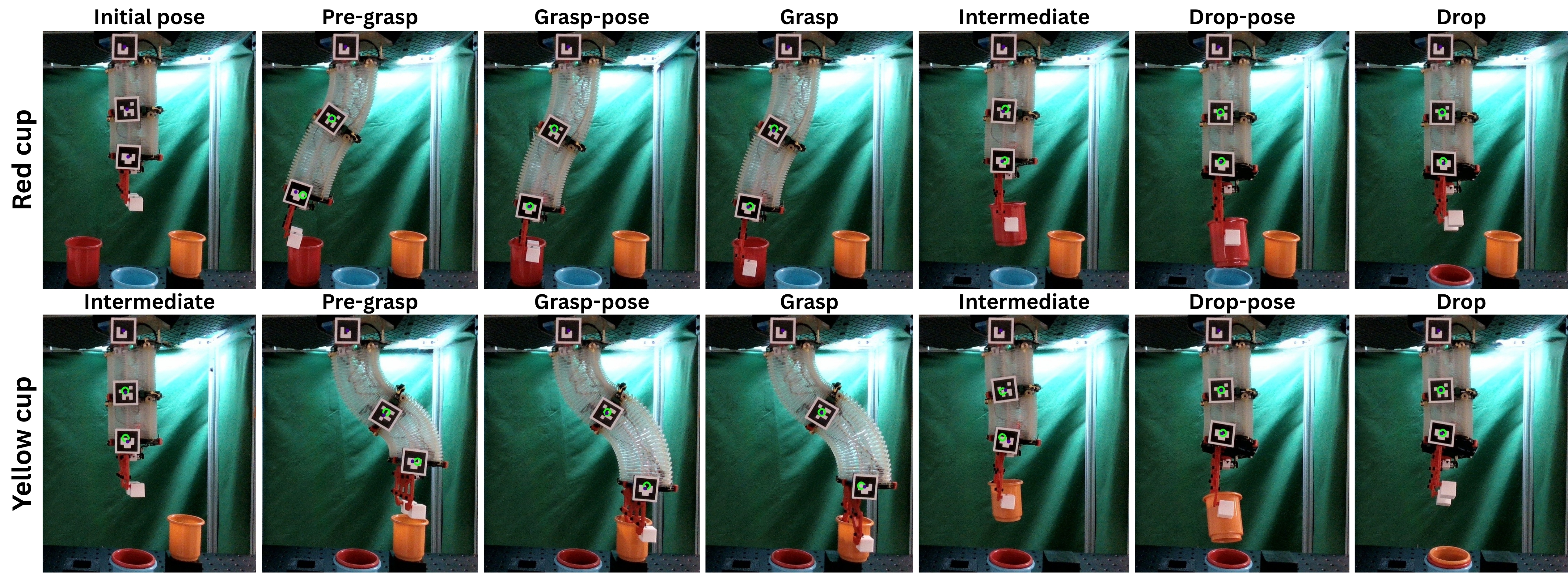}
    \caption[Cup stacking demonstration]{Continuum arm performing the cup stacking demonstration. The red cup is first stacked into the blue cup, and then the orange cup is stacked into the red cup.}
    \label{fig:cup-stacking}
\end{figure}
We utilize stacking cups provided in the YCB object set for manipulation benchmarking~\cite{calliycb}. The task consists of picking and stacking two cups (red and orange) of decreasing size into a larger cup (blue) with a fixed position. A total of $5$ consecutive experiments are performed, out of which $4$ trials are executed successfully. During each trial, the robot servos to pre-designed references that are automatically switched when the control error falls below the threshold. The high success rate for the stacking cups task which requires the robot to precisely achieve the desired shape and end effector pose demonstrates the reliability and accuracy of our controller and its applicability to object manipulation tasks. The series of images in Fig.~\ref{fig:cup-stacking} demonstrate the evolution of a successful trial. Supplementary video S3 shows the full experiment.
\subsubsection{Rice pouring}
\begin{figure}[htbp]
    \centering
    \includegraphics[width=0.6\textwidth]{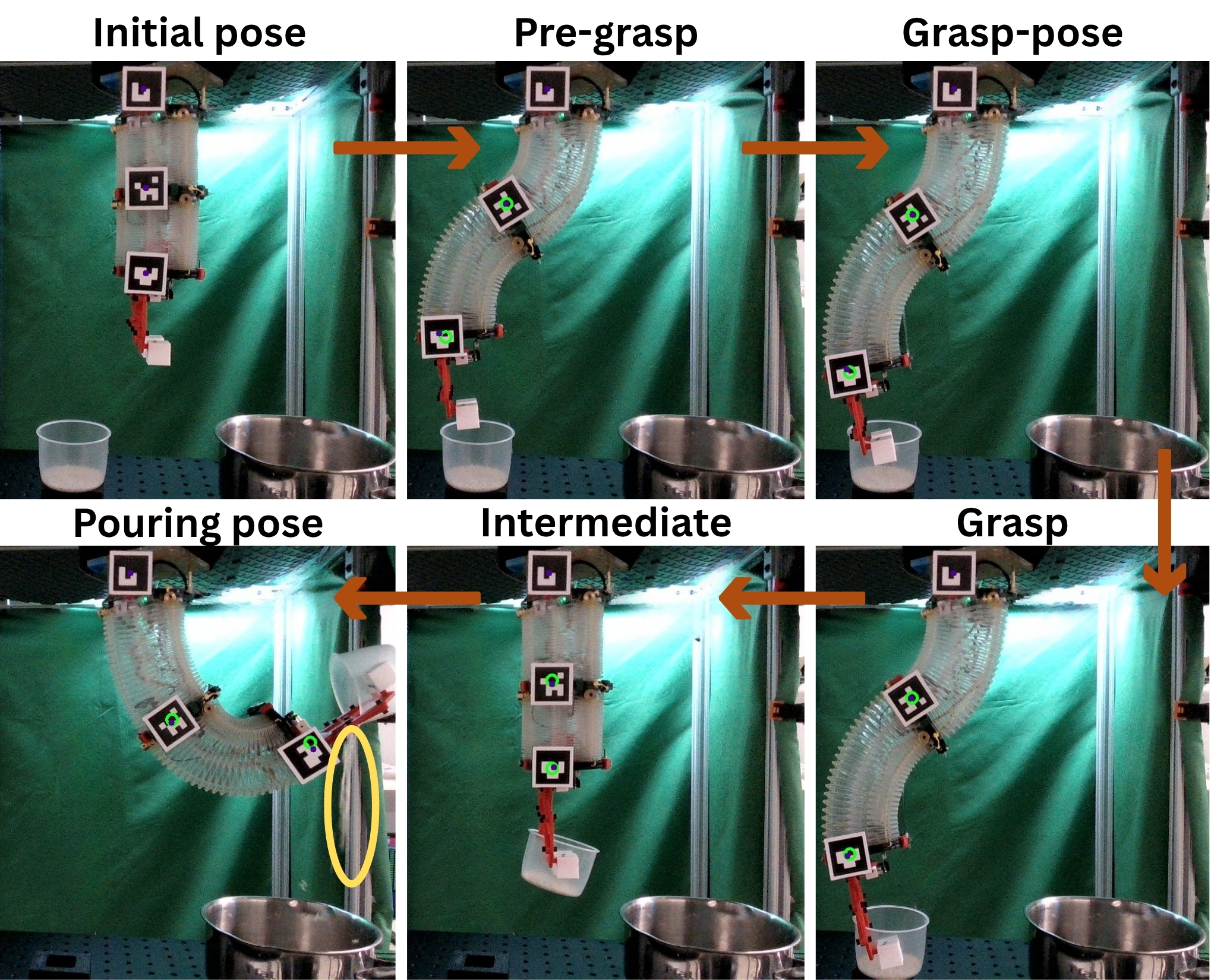}
    \caption[Rice pouring demonstration]{Images demonstrating the evolution of the rice pouring demonstration. The continuum manipulator picks up a measuring cup filled with rice, and maneuvers it to a pouring pose above a steel pot. The rice that is poured out is circled in yellow in the image.}
    \label{fig:rice-pouring}
\end{figure}
A rice pouring task is performed to demonstrate the controller's ability to precisely maneuver grasped objects by regulating the robot's whole-body shape. For this experiment, we utilize a plastic measuring cup filled with 25 grams of rice, and a pot into which the rice is poured. The robot servos to pre-designed references that are switched automatically once the control error drops below the threshold. Images in Fig.~\ref{fig:rice-pouring} show the robot servoing to each of the pre-determined references and successfully pouring rice into the steel pot. The full experiment is presented in supplementary video S4.
\subsubsection{Drawer pulling}
Continuum robots operating in human environments may need to manipulate articulated objects such as drawers. To that extent, we demonstrate a drawer pulling task using our novel shape visual servoing algorithm. Intermediate references are designed to allow the robot to expand, reach for the drawer handle, and shrink to pull the drawer open. Images in Fig.~\ref{fig:drawer-pulling} demonstrate each step and the full experiment is available in supplementary S5.
\begin{figure}[htbp]
    \centering
    \includegraphics[width=0.6\textwidth]{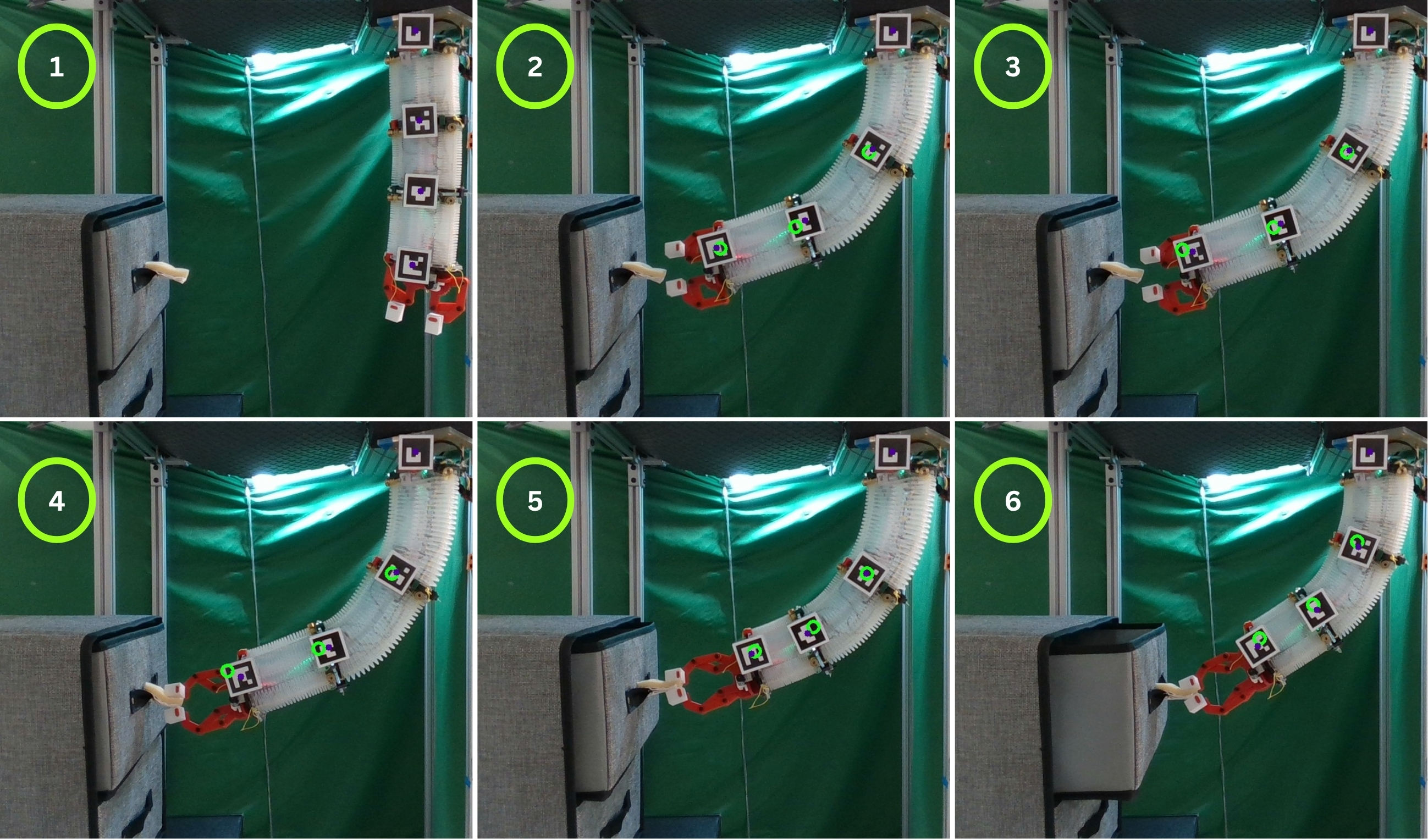}
    \caption[Drawer pulling demonstration]{Image panels 1-6 demonstrating the evolution of the drawer pulling demonstration. The pre-designed shape references enable the continuum manipulator expands to reach for the drawer handle and then shrink in length to pull the drawer open.}
    \label{fig:drawer-pulling}
\end{figure}
\section{Conclusion}\label{sec:conclusion}
This paper introduces a model-based visual servoing framework for controlling the whole-body shape of soft CRs. The proposed shape control approach is beneficial for leveraging the workspace advantages of CRs. A novel robot-shape Jacobian and block diagonal image Jacobian is utilized in the 2-1/2-D visual servoing framework to control the robot's whole-body shape. The Jacobian formulations are modular and as a result the controller can be utilized for CRs with any number of sections. Since the image Jacobian is block diagonal, the individual blocks can be inverted, resulting in faster computation times even with a larger number of image features. The robot=-shape Jacobian itself is sparse and block triangular, which also results in lower computational complexity during inversion operations. Since we utilize the 2-1/2-D visual servoing paradigm the control is globally asymptotically stable. Our experimental results suggest that the controller has high accuracy and is suitable for applications such as object manipulation while operating in cluttered and unstructured environments. However, we noticed that the accuracy of the controller reduces for the three-section robot setup. To improve the accuracy of multi-section continuum arms, investigating the use of controllers that incorporate the system's dynamics should be considered. In the future, our research team is interested in investigating simplified dynamics models and their applicability to whole-body control for soft and CRs. Another advantage of our approach is that we do not require any proprioceptive sensing capabilities on the robot. We estimate the robot's shape purely from the acquired visual feedback, thus simplifying the overall hardware complexity of the system and reducing on-board power requirements. We believe that our work unlocks capabilities that bring us a step closer to utilizing multi-section soft CRs for practical applications.
\section*{Acknowledgments}
The authors would like to acknowledge Neehal Sharma, a graduate student at Worcester Polytechnic Institute, for his indispensable contribution towards building a suitable robotic gripper for the system.
\section*{Authors' Contributions}
A.G.: Robot shape Jacobian design, controller design and development, state estimation design and development (lead), experiment design and implementation, mechatronic system debugging (lead), writing and editing. S.S.C.: robot Jacobian modeling, state estimation (supporting), mechatronic system debugging (supporting). C.D.O.: reviewing and editing (supporting). B.C.: reviewing and editing (lead).   
\section*{Conflict of interest}
No conflicts of interest exist.
\section*{Funding Information}
This work was supported by the National Science Foundation under Award No. 2341532.
\section*{Data access statement}
The datasets generated and analyzed during the current study consist of experimental recordings and control data collected in the laboratory. These materials are stored locally and are available from the corresponding author upon request.
\section*{Ethics statement}
This study did not involve human participants or animal subjects, and therefore no ethical approval was required.
\section*{Supplementary Material}
\begin{itemize}
  \renewcommand\labelitemi{} 
  \item Supplementary video S1
  \item Supplementary video S2
  \item Supplementary video S3
  \item Supplementary video S4
  \item Supplementary video S5
\end{itemize}
\bibliographystyle{ieeetr}
\bibliography{soft_robots_new}
\end{document}